\documentclass[12pt,oneside,reqno,a4paper]{article}
\usepackage[utf8]{inputenc}
\usepackage{amsmath,amsthm}     
\usepackage{lmodern}
\usepackage[T1]{fontenc}        
\usepackage[english]{babel}
\usepackage[numbers]{natbib}    
\usepackage{lipsum}
\usepackage{graphicx}
\usepackage{url}
\usepackage{float}
\usepackage{tabularx}
\usepackage{latexsym}
\usepackage{subfigure}
\usepackage{booktabs}
\usepackage{acronym}
\usepackage{fancyhdr}
\usepackage{subfigure}
\usepackage{subfloat}
\usepackage{titletoc}
\usepackage{dsfont}
\usepackage{enumitem}
\usepackage{caption}

\numberwithin{equation}{section}

\usepackage{multirow}
\usepackage{blindtext}
\numberwithin{equation}{section}

\graphicspath{ {images/} }

\usepackage{ifpdf}
\ifpdf
\pdfoutput=1
\usepackage[bitstream-charter]{mathdesign}
\usepackage[pdfusetitle,colorlinks=true,linktoc=all]{hyperref}
\else
\usepackage[bitstream-charter]{mathdesign}
\usepackage[dvips,ps2pdf,linktoc=all]{hyperref}
\fi
\ifdefined\hypersetup
\hypersetup{
	pdfkeywords={}, linkcolor=black, citecolor=black, filecolor=black, urlcolor=black,
}
\fi
\usepackage{fancyhdr}
\usepackage{blindtext}
\usepackage{fancyhdr}
\usepackage{todonotes}

\title {A comparative study of semi- and self-supervised semantic segmentation of biomedical microscopy data.}

\author{Nastassya Horlava$^1$ \and Alisa Mironenko$^1$ \and Sebastian Niehaus$^{1,2}$ \and Sebastian Wagner$^1$ \and Ingo Roeder$^{1,3}$ \and Nico Scherf$^{1,4}$}

\date{%
    $^1$\small{Institute for Medical Informatics and Biometry, Carl Gustav Carus Faculty of Medicine, TU Dresden, Germany}\\%
    $^2$\small{AICURA medical GmbH, Berlin, Germany}\\%
    $^3$\small{National Center of Tumor Diseases (NCT),  Partner Site Dresden, Germany}\\%
    $^4$\small{Max Planck Institute for Human Cognitive and Brain Sciences, Leipzig, Germany}\\%
}

\begin{document}
\maketitle	
	
\begin{abstract}
In recent years, Convolutional Neural Networks (CNNs) have become the state-of-the-art method for biomedical image analysis. However, these networks are usually trained in a supervised manner,  requiring large amounts of labelled training data. These labelled data sets are often difficult to acquire in the biomedical domain. In this work, we validate alternative ways to train CNNs with fewer labels for biomedical image segmentation using. We adapt two semi- and self-supervised image classification methods and analyse their performance for semantic segmentation of biomedical microscopy images.
\end{abstract}
	
	\renewcommand{\thepage}{C-\Roman{page}}
	
	\pagenumbering{arabic}

	\section{Introduction}
	\markboth{Introduction}{}
	In recent years, supervised machine learning approaches showed spectacular results in various image analysis problems \cite{LeCun15}. Based on massive, annotated data sets, deep learning systems have come to the point where they are on par or even outperform humans in specific tasks \cite{Silver18} \cite{Hannun2019} \cite{Esteva2017}. However, fully annotated data sets are typically not available or even feasible to create in many domains. Manual reference annotations for pixel-level semantic segmentation in biomedical imaging are particularly costly as they can be too time-consuming and require considerable expert knowledge that might not readily be available. Here, semi- and self-supervised learning methods are promising approaches to build generalizable segmentation tools as they can leverage raw data and require only a few or no labels at all. These methods yield encouraging results in computer vision tasks on natural images. 
	Biomedical image data however, differs in a few key aspects from natural images. In particular biomedical images typically exhibit:
	\begin{itemize}
	    \item less discriminative textural features,
        \item lower signal-to-noise ration,
	    \item a high degree of structural self-similarity (e.g. organic shapes), and 
	    \item repetitive structures (e.g. cells).
	\end{itemize}
	While some of these specific properties of biomedical images can make segmentation tasks more complicated, some aspects like self-similarity might help Machine Learning approaches to learn from fewer examples.
	Consequently, the goal of this work is to explore how prototypic methods of semi- and self-supervised image classification from the field of pure computer vision perform in the task of semantic segmentation (pixel-level classification) for biomedical microscopy data. In this comparison, we will focus on segmentation accuracy, as well as the data- and label-efficiency. 
	
	We first explore a semi-supervised approach where we work with only partially labelled data using Unsupervised Data Augmentation (UDA) \cite{Xie2019} and study the influence of a varying number of labelled training data on segmentation accuracy. In some application cases, however, we have to work with datasets where we have no labels at all. This typically happens when we deal with data from novel imaging modalities, new model systems or exploratory bioimaging studies where specific expert knowledge is required to annotate

	\section{Methods}
	\markboth{Methods}{}
	
	\subsection{Datasets}
	
	We tested these methods on openly available microscopy datasets. In this work, we focus on the two major application domains we are interested in: cell-level analysis in developmental and cell biology and tissue-level analysis in computational neuroanatomy. Thus, we chose the following datasets for our initial tests:
	\begin{itemize}
		\setlength\itemsep{0cm}
		\item	\emph{PhC}: The PhC-C2DH-U373 (further in the text called \emph{PhC}) dataset showing cells from the Glioblastoma-astrocytoma U373 line in phase-contrast microscopy from \cite{Maska2014}, \cite{Ulman2017}
		\item	\emph{FluoGFP}: The Fluo-N2DL-HeLa dataset from the MITOCHECK project \cite{Maska2014} \cite{Ulman2017} showing the fluorescence signal of HeLa cells stably expressing a Histone H2b-GFP reporter,
		\item \emph{FluoHoechst}: The BBBC039 (further called Fluo-Hoechst) dataset from \cite{Ljosa2012} showing the fluorescence signal of the DNA channel (Hoechst) of U2OS cells,
		\item \emph{BigBrain}: As an example of a challenging tissue-level semantic segmentation problem, we use the BigBrain dataset from \cite{Amunts2013} consisting of histological sections of a human brain imaged with light microscopy. 
		
	\end{itemize}
	
	\subsection{Preprocessing of image data}
		For the PhC, FluoGFP, and FluoHoechst we were interested in the binary segmentation task (cells vs background). The images from the different datasets were all cropped to the same square 512*512 size. Further details of the datasets are summarized in Table 1. 
		
		For the BigBrain dataset, we generated 2D training data from the high-resolution 3D microscopy scans (100um isotropic (1970x2330x1890) by randomly rotating the scan and taking the 2D slices as images for the training. The dataset contained nine different classes. For this work, we focussed only on the problem of segmenting white and grey matter, thus restricting ourselves to segmenting only these classes (original classes 2, 3, 5), while considering pixels of all remaining classes as background. We cropped the images to patches of 600*600 pixels for the semi-supervised approach to patches of 256*256 for the self-supervised case.

	\begin{table}[H]
		\centering
		\caption{Overview of between different datasets
			\label{methods}} \vspace*{2mm}
		\begin{tabular}{| m{2.2cm}| m{1.5 cm}| m{2cm} | m{2.3cm}| m {2.3cm}| m{2.5cm}| }
			
			\hline
			Dataset & Number of samples  &	Original size of the image  &	Cropped size of the image \break (semi-supervised) &	Cropped size \break  (fully self-supervised) &	Number of classes  \\ \hline
			PhC& 114& varies& 512*512 & 128*128& 2 \\ \hline
			FluoGFP&	92&	varies&	512*512&	128*128&	2 \\ \hline
			FluoHoechst &	200&	varies&	512*512&	128*128&	2 \\ \hline
			BigBrain &	3479  &	770*605  &	600*600	 & 256*256  &	9 originally, \break 4 after preprocessing \\ \hline

		\end{tabular}
	\end{table}
	
	For each dataset, we normalized the training and testing data according to the maximum and minimum values of the training data. As data augmentation is specific to the particular learning approach, and we thus performed different processing for self-supervised and semi-supervised methods.
	
	\subsection{Data Augmentation and Training}
	
	\subsubsection{Semi-supervised Learning}
	
	As a semi-supervised approach, we adapted the Unsupervised Data Augmentation method \cite{Xie2019}. In the paper, the authors of UDA focus on image classification tasks. Here, we extend this idea to the task of semantic segmentation: 
	
	\begin{itemize}
	    \item We used a U-net architecture \cite{Ronneberger2015} as a suitable backbone for semantic segmentation tasks,
	    \item We applied spatial augmentations to the ground-truth mask alongside the corresponding image,
	    \item We introduced an additional step to bring the predicted masks of the image, and the mask of it's strongly augmented version to the same spatial position before calculating the unsupervised loss.
	\end{itemize}
	
	The main idea of the UDA method is to use both labelled and unlabeled samples simultaneously during the training. In practice, we had a fully labelled dataset, from which we then constructed a labelled subset by randomly drawing sample images with their masks from the full dataset. We then created the unsupervised subset by taking the samples that were left from the full dataset without their corresponding segmentation masks.
	
	\textbf{Training pipeline.} 
	Within one epoch, we do the following: 

	Each pair (image, mask) from the supervised subset is first augmented using the set of \emph{weak} transformations (see the section on Data Augmentation for details), and then the supervised loss is computed between the predicted mask and the original ground truth one. In this work, we use the Dice loss \cite{Milletari2016} as a supervised loss as the classical choice for dense segmentation tasks \cite{Sudre2017}. 
	\begin{figure}[h]
		\centering
		\includegraphics[width=\linewidth]{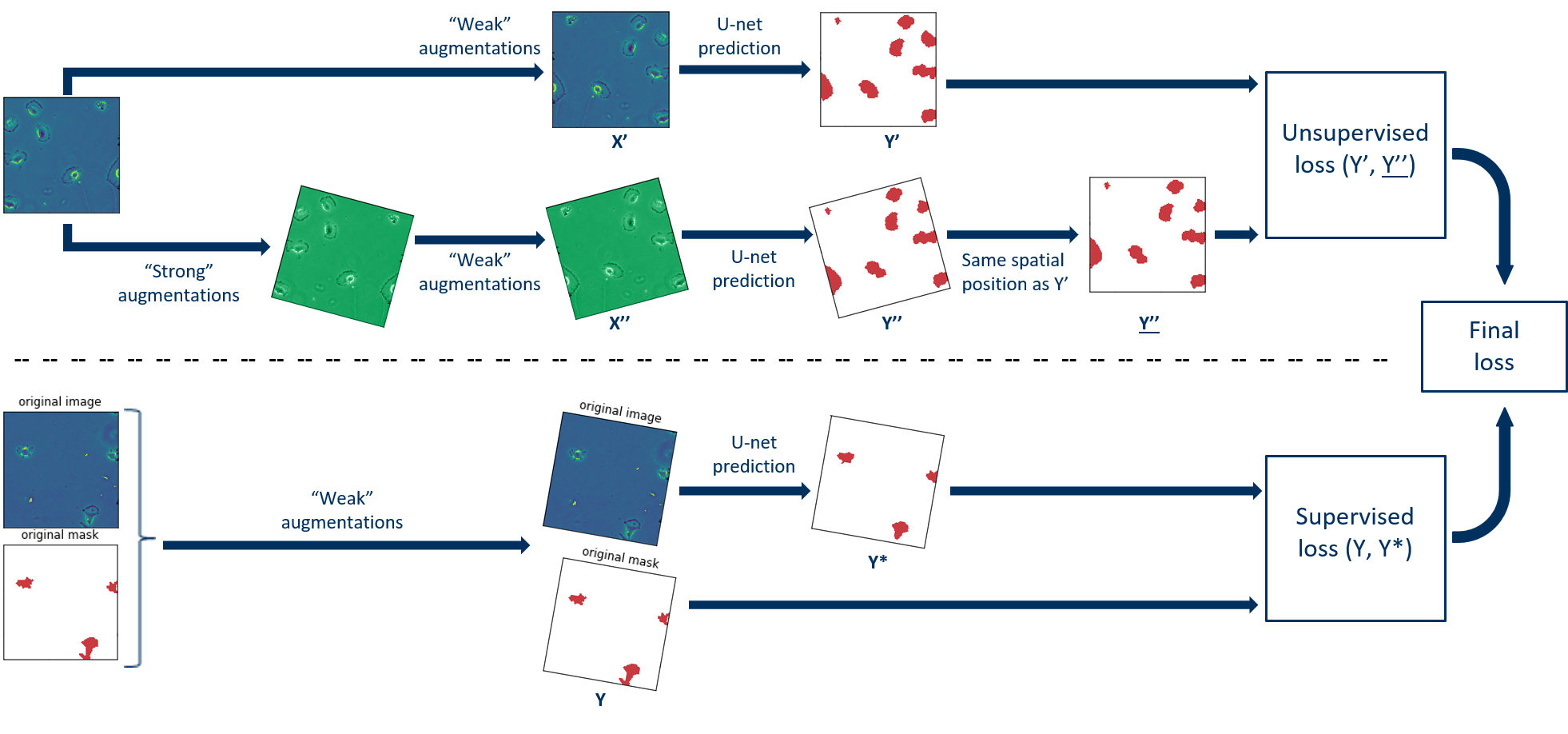}
		\caption{General training pipeline within one epoch. The upper half of the figure represents the unsupervised part for one image, while the bottom part - supervised part for 1 image and its mask.}
		\label{fig:1}
	\end{figure}

	An image X from the unsupervised subset is first augmented using a random subset of \emph{strong} transformations (see section on Data Augmentation for details), resulting in an augmented unlabelled image X’’. The same set of \emph{weak} transformations are then applied to an unlabelled image and its augmented version. The unsupervised loss is then calculated between the masks, predicted for the unlabeled image and its augmented version ( Y’ and Y’’ respectively), following the idea that produced masks should be similar since they initially represent the same image. Note that if the affine transformations were applied when creating the augmented image X’’, then the predicted mask of the augmented sample Y’’ is brought to the same spatial position as the predicted mask of the weakly transformed, unlabeled image Y’ before calculating the unsupervised loss. We align the images by applying the inverse affine transformations to the mask of the augmented sample Y’’. For the unsupervised loss, we compared two different versions: Kullback–Leibler divergence \cite{Kullback} (later in the text called KLDiv), and the IIC loss \cite{Ji2019} extended to the task of dense prediction (later in the text called IID).
	
	We compute the final loss for the backpropagation as the sum of supervised and unsupervised losses. The general overview of this approach is sketched in Figure \ref{fig:1}.

	One of our major aims was to explore the feasibility of semi- and self-supervised learning in the biomedical domain where we only have sparse labels (e.g. a small percentage of the available images are labelled). Thus, we wanted to see how the algorithm performed with different ratios of labelled samples. The amount of labelled and unlabeled data was therefore usually unequal leading to unbalanced training. To alleviate this technical problem, We loop over the smaller subset while using the same batch size for both labelled and unlabeled parts in each epoch. 
	
	To avoid overfitting, we saved the network state that achieved the lowest loss for the validation set. Considering that we calculated three different losses, we found meaningful to save either the state of the network that yields the best supervised loss, or the state that yields the best final loss. We will then compare the results of these two states.

	\textbf{Data augmentation.} For the training phase, 2 different subsets of augmentations were created: \emph{weak} and \emph{strong} set of transformations. 
	The \emph{weak} set consisted of the following transformation:
	\begin{enumerate}
		\item	Scaling to a random ratio between 95\% and 105\% of original size of the image,
		\item	Rotation for a random angle between -15$^\circ$ and 15$^\circ$, and
		\item	Horizontal flipping with 50\% probability.
	\end{enumerate}
	
	We adapted the \emph{strong} set of transformations from the RandAugment method \cite{Cubuk2020}, which was also used in \cite{Xie2019} for the image recognition task. RandAugment is a simplified version of AutoAugment method \cite{Cubuk2019}, but unlike AutoAugment, it does not require searching for the most optimal augmentation policies, thus saving both computational resources and number of samples needed for training.
	
	In RandAugment, for each augmentation policy, we randomly sample transformations from the full set of 9 different transformations. These transformations can be technically divided into two groups: \emph{spatial} (affine) transformations, which transform the image, and to which an explicit inverse transformation exists, and \emph{non-spatial} transformations, later called \emph{general transformations} that change  only the intensity of pixels, but not their spatial position. These general transformations do not require a reverse transformation of the resulting segmentation masks.
	
	From the affine transformations, we used only a rotation with a random angle between 0 $^\circ$ and 360$^\circ$ . 
	
	The general transformations included the following:
	
	\begin{enumerate}
		\item	AutoContrast, which maximizes the global image contrast,
		\item	Invert, which inverts the pixel values for all channels,
		\item	Equalize, which performs histogram equalization,
		\item	Solarization, which inverts pixel values above a certain threshold,
		\item	Contrast,which adjusts the contrast of the image,
		\item	Color, which adjusts the colorization level of the image,
		\item	Brightness, which changes the brightness level, and
		\item	Sharpness, which either sharpens or blurs the image.
	\end{enumerate}
	
	The effect of applying these general transformations to biomedical images is demonstrated in Figure \ref{fig:2} using the PhC data. 
	\begin{figure} [h]
		\centering
		\includegraphics[width=0.6\linewidth]{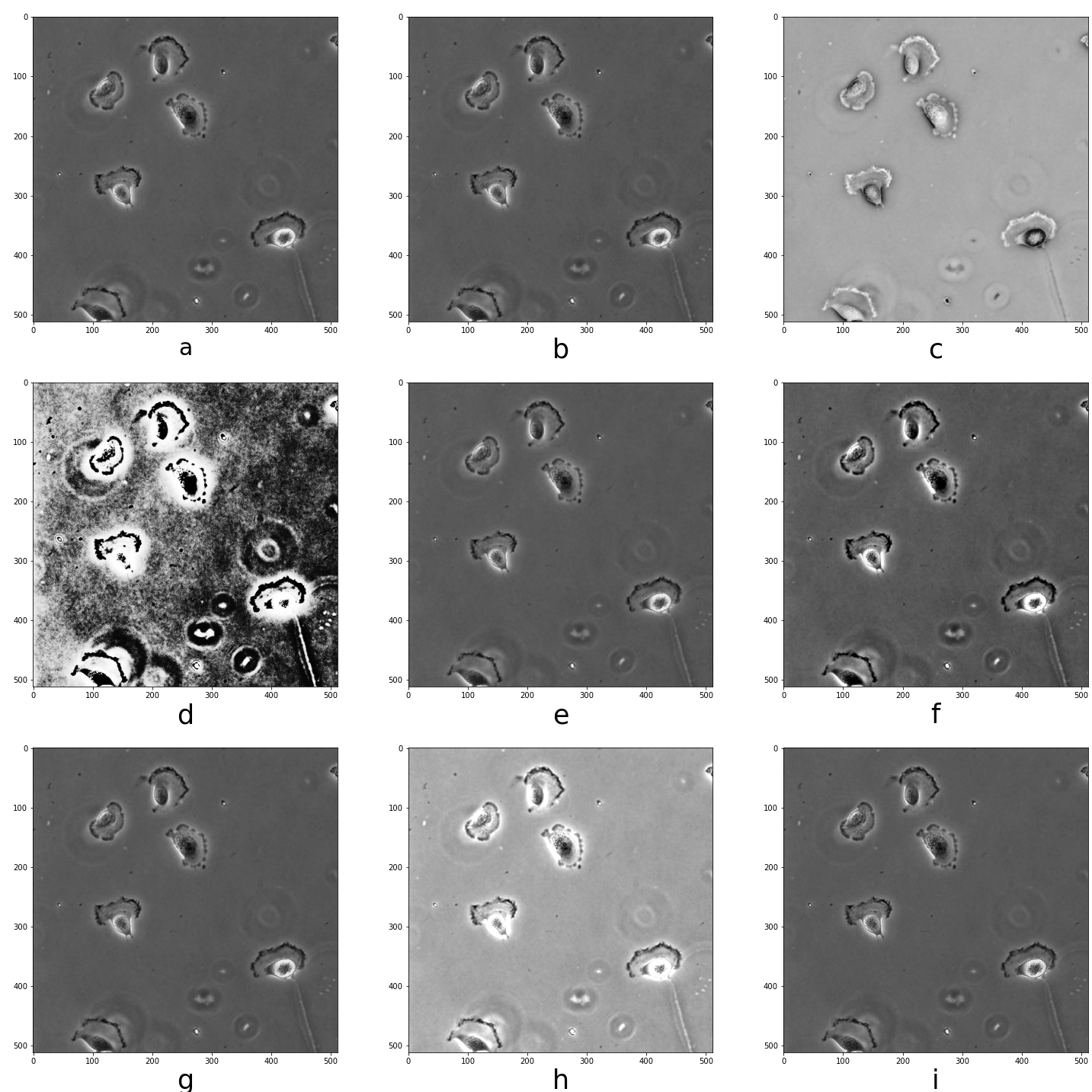}
		\caption{Example of the image after applying to it one of the 8 general transformations with magnitude 30 out of 30 where applicable. (a) original image (b) AutoContrast (c) Invert (d) Equalize (e) Solarization (f) Contrast (g) Color (h) Brightness (i) Sharpness.}
		\label{fig:2}
	\end{figure}

	\subsubsection{Self-supervised approach}
	
    In this work, we focus on Invariant Information Segmentation \cite{Ji2019} as a self-supervised approach that does not require any reference labels at all. The core idea of this method is to maximize the mutual information between automatically generated segmentation masks of paired images that are obtained from the same source image by applying different augmentations. As the augmentation policy, the rotation for a random angle between 0$^\circ$ and 360$^\circ$ was used. That saying, during training, we minimized the negative mutual information between the segmentation mask of the image and its rotated version. It should be noted that to apply this method, we have to define the number of classes beforehand, which could introduce an unwanted bias for noisy data or data with multiple highly unbalanced classes. Following the ideas of the original paper, we also introduced an additional segmentation head with an increased number of classes in order to handle noisy data.
	
	\subsection{Implementation details}
	For both approaches (UDA and IIC), we used the original U-net as proposed in \cite{Ronneberger2015}. This is one of the most widely used architectures in dense segmentation tasks, especially for biomedical applications. Further technical details of the training setup for both approaches are summarized in Table 2.
	
	\begin{table}[H]
		\centering
		\caption{Details of the training setups for self-supervised and supervised approaches.
			\label{methods}} \vspace*{5mm}
		\begin{tabular}{ m{4.2cm}| m{4.2 cm}| m{4.2cm} }
			
			Parameter  &	Semi-supervised approach UDA & Self-supervised approach IIC \\ \hline
			Optimizer&	Adam &	RMSProp\\ \hline
			Learning rate&	0.001 with step-based learning rate schedule&	0.01 \\ \hline
			Number of epochs &	100&	10 \\ \hline
			Batch size&	1&	10 \\ \hline		
		\end{tabular}
	\end{table}

	\section{Experiments and Results}
	\markboth{Results}{}
	\subsection{Semi-supervised approach}
	
	For the semi-supervised approach, we compared the following parameters of the training setup for each of the datasets:
	\begin{enumerate}
		\item	The ratio of labeled dataset: either 10,
		25 or 50\% of the full labels,
		\item	The unsupervised loss: either KLDiv or IID,
		\item	The saved network state: either based on the best supervised loss or on the best final loss.
	\end{enumerate}

	\begin{figure} [h]
		\label{f3}
		\centering     
		\subfigure[]{\label{fig:a}\includegraphics[width=0.49\linewidth]{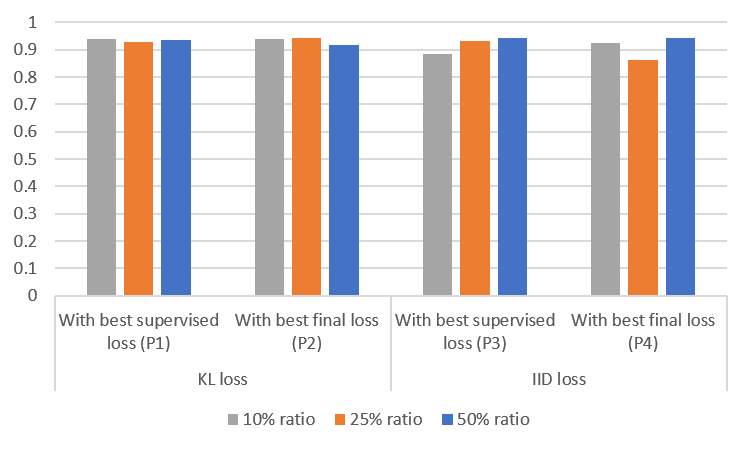}}
		\subfigure[]{\label{fig:b}\includegraphics[width=0.49\linewidth]{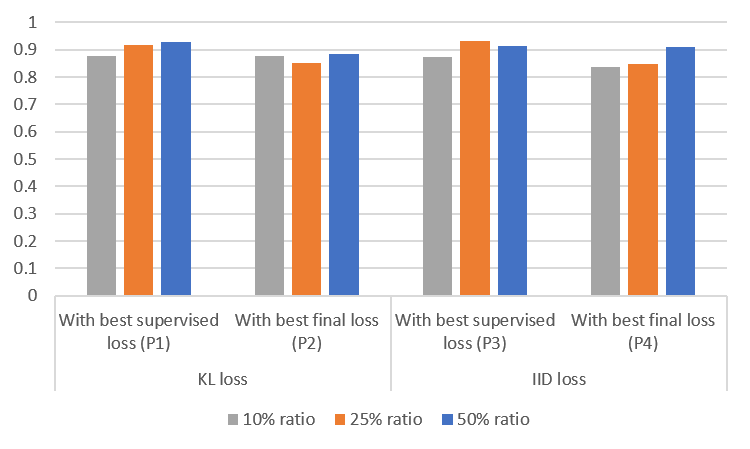}}
		\subfigure[]{\label{fig:c}\includegraphics[width=0.49\linewidth]{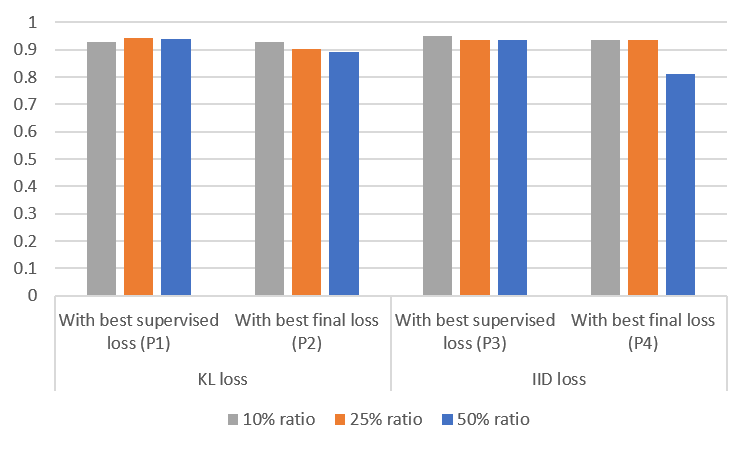}}
		\subfigure[]{\label{fig:d}\includegraphics[width=0.49\linewidth]{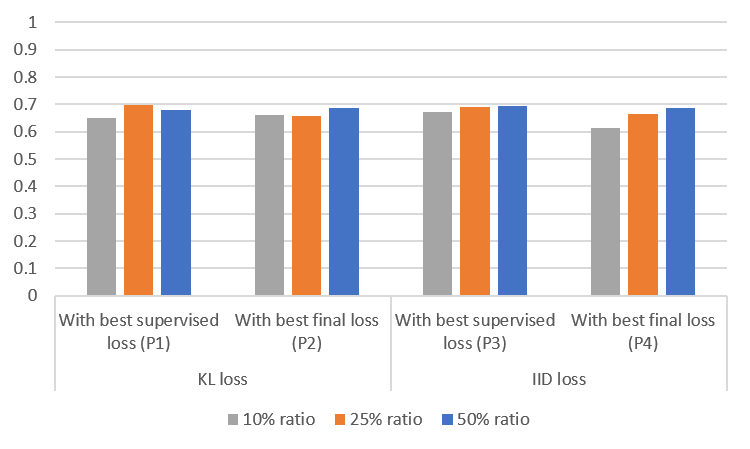}}
		\caption{Barcharts representing the mean IoU values, obtained withing different training pipelines(P1/P2/P3/P4) for different ratios of the labeled data. The following datasets are presented: (a) PhC dataset, (b) FluoGFP dataset, (c) FluoHoechst (d) BigBrain dataset}
	\end{figure}

	Among all four datasets network states with the best supervised loss slightly outperformed the one with the best final loss. 

	As a quantitative metric for segmentation accuracy, we use the well-established mean Jaccard similarity metric (also called IoU) that we calculated for the test subset of the
	\begin{figure} [H]
		\centering     
		\subfigure[]{\label{fig:a}\includegraphics[width=0.48\linewidth]{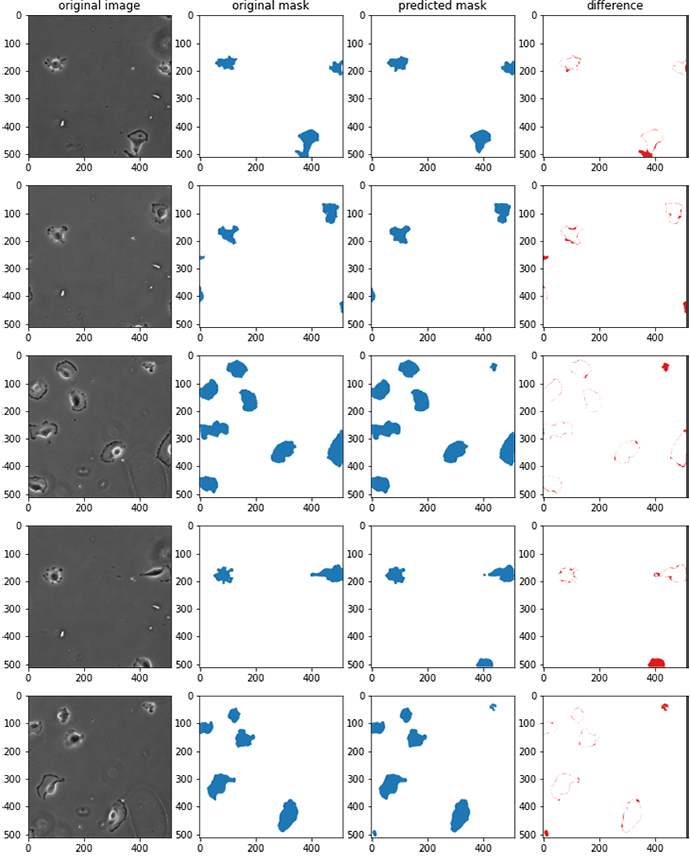}}
		\subfigure[]{\label{fig:b}\includegraphics[width=0.48\linewidth]{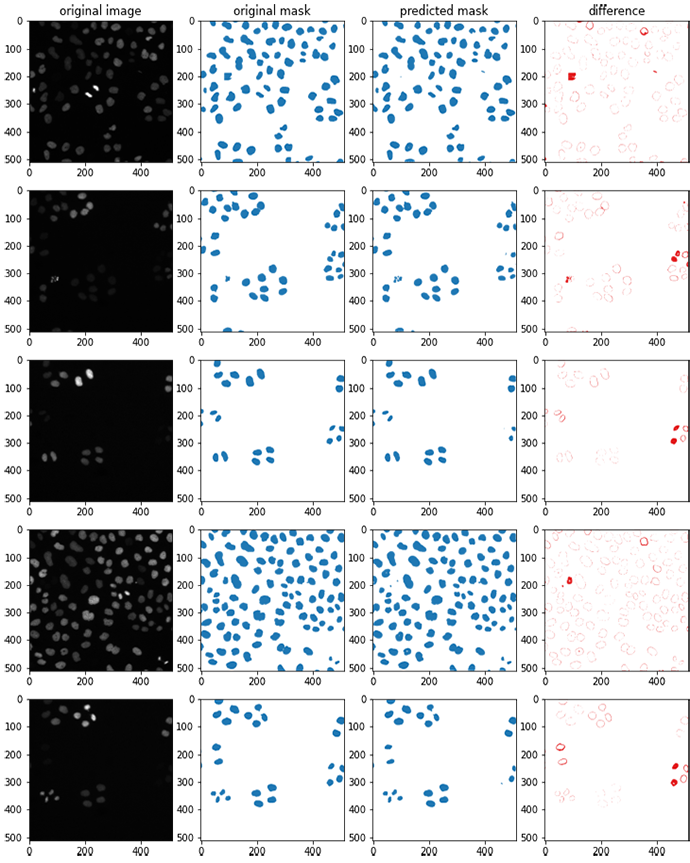}}
		\subfigure[]{\label{fig:a}\includegraphics[width=0.48\linewidth]{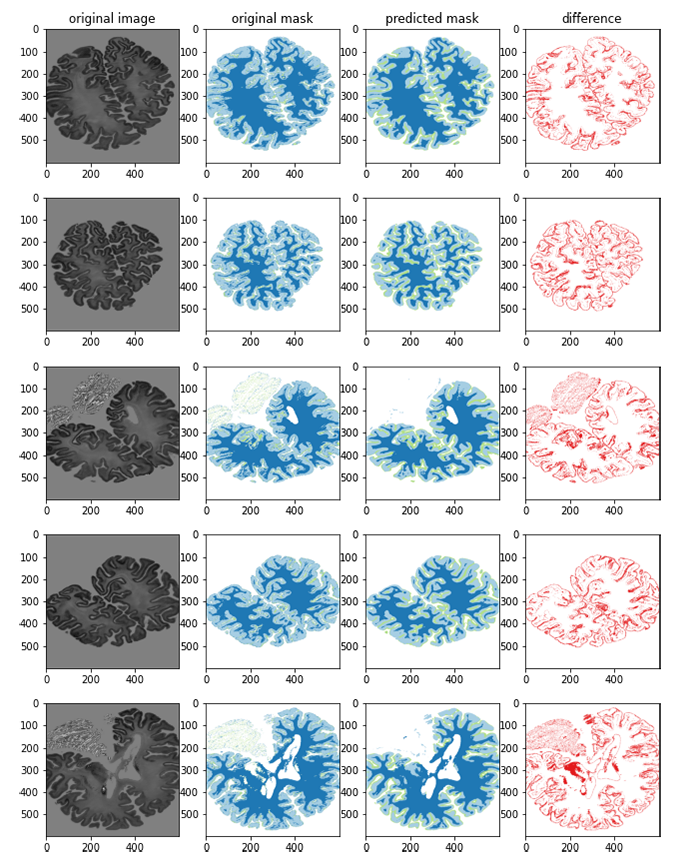}}
		\subfigure[]{\label{fig:b}\includegraphics[width=0.48\linewidth]{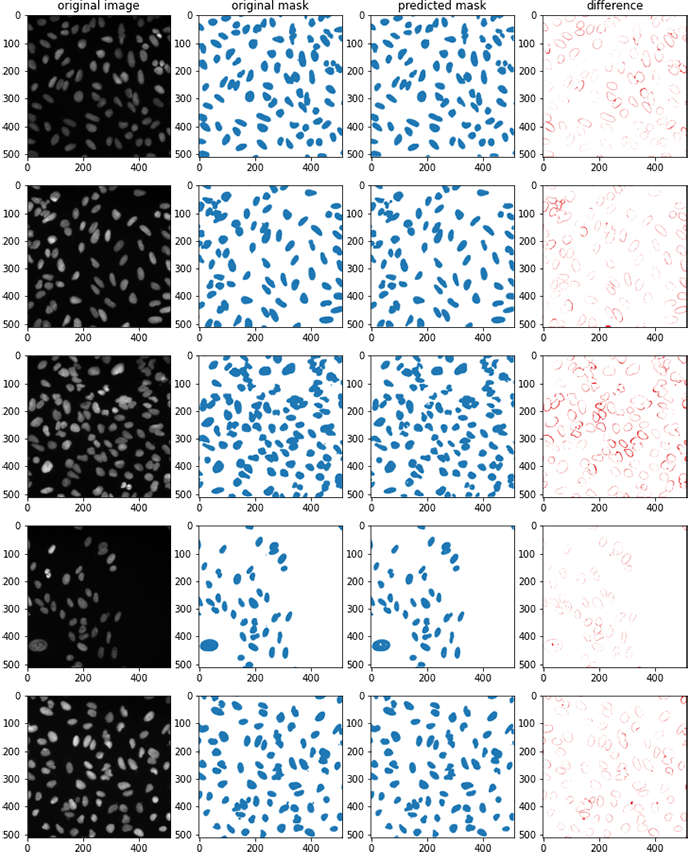}}
		\caption{Example of segmentation with the semi-supervised UDA approach. Here, the results are presented for the pipeline where 25\% of the labeled data was used, IID loss as unsupervised loss and the state of the network with the best supervised loss. Original image (column 1), it’s ground truth(GT) segmentation masks (column 2), predicted mask (column 3) and difference between GT and predicted masks (column 4) for the following datasets: (a) PhC dataset, (b) FluoGFP dataset,
		(c) BigBrain dataset (d) FluoHoechst.}
		\label{fig:f4}
	\end{figure}
	data. For two matching objects A (result) and B (reference), the IoU is calculated as follows:
	\begin{equation*}
		IoU=\frac{|A \bigcap B|}{|A \bigcup B|}
	\end{equation*} 
	
	The obtained mean IoU for all datasets for all types of training pipelines are presented in Figure 3. The exact numerical values can be found in Table S1 in Supplementary Material. 

	Somewhat surprisingly, we do not see major effects of the different label ratios. We observed some minor trends  within the same training pipeline varied for different datasets:
	\begin{itemize}
		\item	For PhC dataset, the positive correlation between the number of labelled examples and obtained accuracies was observed only when the IID loss was used during the training and network state with best supervised loss was used (P3). In other training pipelines, the IoU varied differently and we found no connection with the ratio. 
		\item	For FluoGFP dataset, the positive correlation was observed in pipeline P1 and P4
		\item	For Bigbrain dataset, this correlation was observed in P3 and P4 pipelines.
	\end{itemize}
	
	Concerning the various datasets, the results achieved in for tissue segmentation the BigBrain microscopy dataset were significantly worse than those from the cell segmentation problems in the fluorescence (FluoGFP, PhC) and phase-contrast (FluoHoechst) microscopy datasets. At the same time, we observed no significant difference in performance was observed between FluoGFP, PhC, and FluoHoechst datasets.

	Visual examples of the segmentation results are shown in Figure 4. Since there are too many different setups to show all results, we focus on the results of P3 pipeline for 25\% of the data are presented for each of the datasets here. 
	
	\subsection{Self-supervised approach}
	
	The obtained mean IoU for all datasets for both main and auxiliary heads are presented in Figure 5. Note that for the auxiliary output, classes were assigned in accordance to maximally overlapping areas with ground-truth labels. The exact numerical values can be found in Table S2 in the Supplementary Material. 
	
		\begin{figure}[h]
		\centering
		\includegraphics[width=0.55 \linewidth]{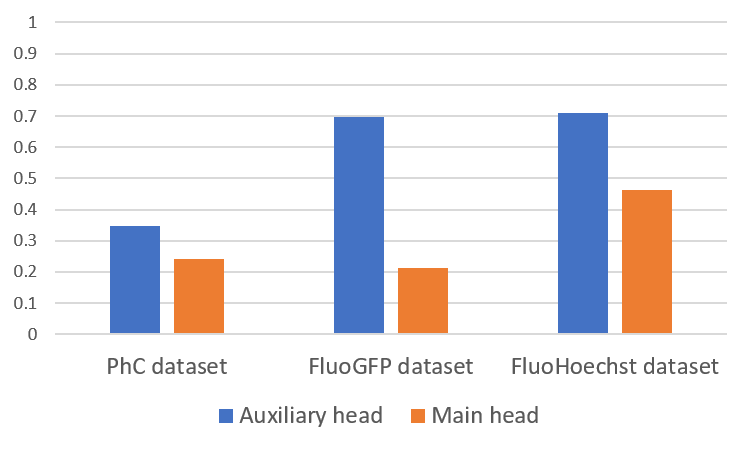}
		\caption{ Barcharts representing the mean IoU values for different datasets, obtained for the output of the main and auxiliary head in the self-supervised approach. }
		\label{fig:2}
	\end{figure}
	
	The visual representation of segmentation via the self-supervised approach is shown in Figure 6. 
	The resulting segmentation masks for the Big Brain dataset suggest that this self-supervised method is not suitable for the semantic segmentation of tissue structures. Although the network was able to capture the general structure of the brain, it was not able to distinguish internal tissue substructures. Consequently, for the simpler task of binary segmentation into brain and background, the results are much better.
	
	The situation is similar in cases of cell images. In certain situations (Fig. 6c-e and Fig. S1) the respective neural networks were able to find meaningful patterns but was typically unable to find semantically correct masks in the binary segmentation case. The cell tracking datasets also reveal that this method (having no training information on how an object is supposed to look like) is essentially concentrating on average brightness differences and is thus very sensitive to noise and structures in the background. The fact that in case of FluoHoechst dataset (Fig. 6e) the auxiliary segmentation head works better than the main one may be because the method could also sensitive to the noise imposed by the data retrieval method, which the auxiliary head then identifies as a separate class.
	
			\begin{figure}[H]
		{     
			\subfigure[]{\label{fig:a}\includegraphics[width=0.85\linewidth]{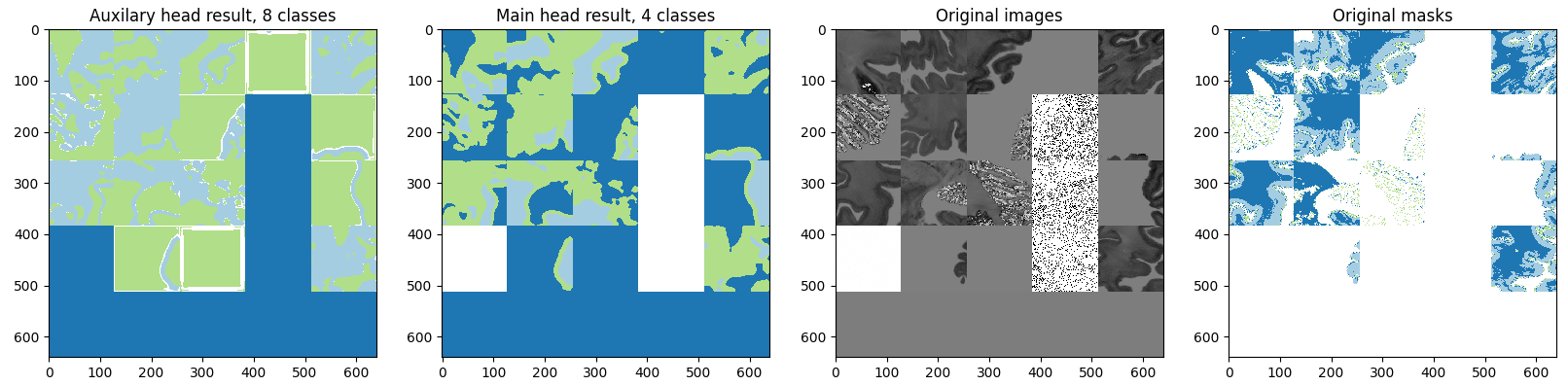}}
			\subfigure[]{\label{fig:b}\includegraphics[width=0.85\linewidth]{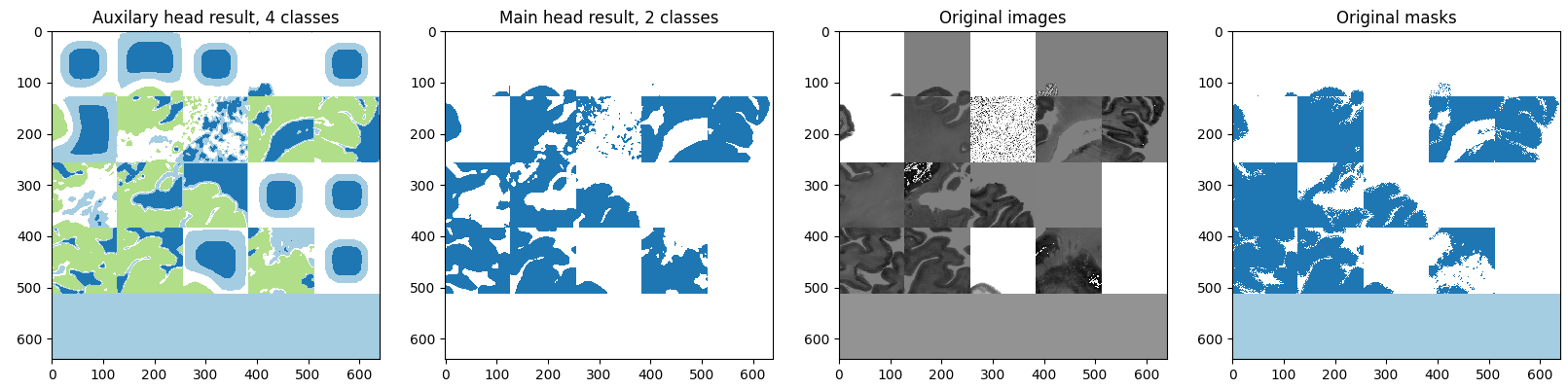}}
			\subfigure[]{\label{fig:c}\includegraphics[width=0.85\linewidth]{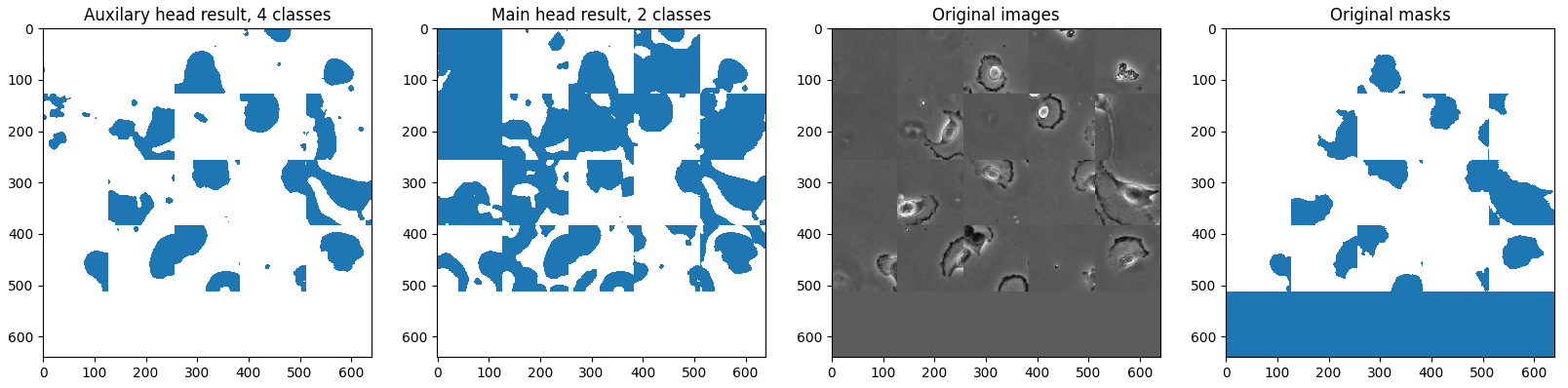}}
			\subfigure[]{\label{fig:e}\includegraphics[width=0.85\linewidth]{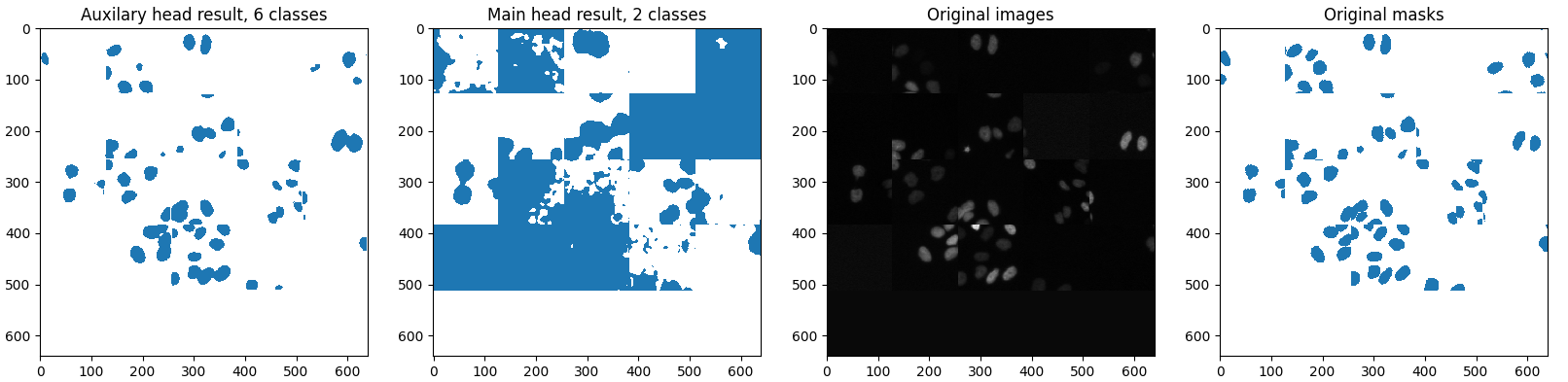}}	
			\subfigure[]{\label{fig:f}\includegraphics[width=0.85\linewidth]{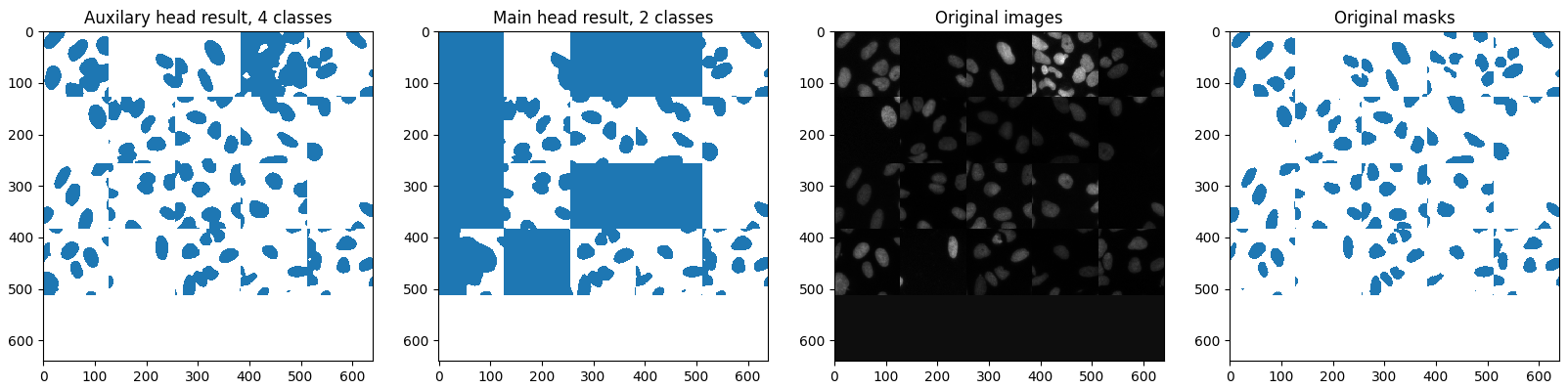}}
			\caption{Example of segmentation from the self-supervised IIC approach. Predicted mask from auxiliary head (column 1),predicted mask from the main head (column 2), ground truth segmentation masks(column 3), and original image(column 4) for the following datasets: (a) BigBrain dataset with 4 classes (b) Big Brain with only background detection; the average binary cross entropy loss for the whole dataset is 0.23 (c) PhC dataset (d) FluoGFP dataset (e) FluoHoechst }}
		\label{fig:f5}
	\end{figure}
	\section{Discussion}
	\markboth{Discussion}{}
	
	Given the preliminary results obtained for the different ratios of labelled data in semi-supervised approach, we cannot conclude that there is a clear effect of labelled data size to the obtained accuracy above 10\% of labels. However, the fact that the accuracy does not drop dramatically with a smaller number of labelled samples suggests that a semi-supervised approach performs consistently well and could be of use even in the tasks with small amounts of labelled data. Regarding the efficiency, we found no indication that training times differ much between the setups (different pipelines and ratios of labelled data in the semi-supervised approach). As the next step, we will consider label ratios below 10\% and explore this low label regime in more detail. 

	
	As for the fully self-supervised approach, it performed poorly for most of the tasks except for simple binary segmentation tasks where differences in image intensity are consistent between the classes. This suggests that semantic segmentation of biomedical images is a difficult task for this type of algorithm alone. That is probably due to background structures and noise and the complexity of biological structures to be segmented (e.g. thin tissue layers in the sulci of cortical folds) and the lack of training labels, i.e. there is no information for the network on how the structures of interest should look like at training time. That being said, for the simple tasks, self-supervised approaches still could be useful as part of an analysis pipeline to provide intermediate results that can then be filtered and processed. Instead of providing the desired number of classes exactly, it seems more promising to aim for an oversegmentation (in terms of classes) and filter the results later on (compare the 4-class and 2-class output in Fig. 6e).
	 We also think that self-supervised methods are exciting candidates to bootstrap an active-learning approach where subsequent feedback from a human annotator could be used in an adaptive, semi-supervised solution. 

	A problem we ran into with self-supervised approaches was that it can be difficult to asses the performance of the network based solely on the loss function value. The smaller loss does not necessary indicate that network is solving the correct task: for example, when looking for the optimal setup, we have seen cases when network converges to a degenerate solution of ignoring the input and always predicting the same output. In such cases, the loss function value was also sometimes smaller than the loss of the network state with more meaningful results. This leaves the further question of how long to train the self-supervised algorithm and when to stop to avoid such cases.
	
	There are still some challenges in learning how to use the full potential of the unlabeled data for biomedical segmentation. It still remains an open question in which cases (or problem domains) meaningful results could be obtained without any labelled data in complex tasks such as semantic segmentation. As our study suggests, generic self-supervised approach such as \cite{Ji2019} is not yet able to achieve results comparable to fully supervised or semi-supervised methods in biomedical image segmentation tasks like brain substructures segmentation. One way to tackle this problem would be to develop more specialized models that introduce an inductive bias for specific kinds of structures to be segmented. On the contrary, if classes represent simple structures that show a consistent difference in appearance (in particular image intensities), generic self-supervised approaches still could be of use. 
	
		A clear direction for future research is to explore if we can use previously labelled datasets as a pretraining step to segment new unlabeled ones in a transfer learning setting. In a situation where one is given a fully unlabeled dataset of some structure, i.e. tissue structure in the brain, it would be interesting to investigate whether it is possible to combine the transfer learning approach with the fully unsupervised one. That is, to use the previously pre-trained on different brain images network and refine it to the new dataset via a fully self-supervised approach.

	Semi-supervised approach, on the other hand, shows a clear potential for biomedical image segmentation. As our study showed, even a small amount of the labelled data would be efficient enough to obtain results, comparable to fully supervised approaches. And with the help of tools such as LabKit \cite{LabKit} or QuPath \cite{Bankhead2017}, a small amount of the data can be labelled in many applications like cell segmentation. For more complex labelling tasks like in the case of neuroanatomy where a considerable degree of expert knowledge is needed, we might need more efficient ways to use a semi-supervised approach, e.g. in an active learning setting.

	One important limitation of our current study is that we focused on 2D images datasets and treated 3D datasets as a collection of independent 2D slices. Yet there exists plenty of 3D or even 3D+time datasets in biology, for which the consistency of structures along the spatial or temporal dimension could give attractive targets for self-supervised methods. Investigating the ways of using such information posts an essential direction of further investigations 
	
	\newpage
	\bibliographystyle{unsrt}
	\bibliography{references}
	\addcontentsline{toc}{section}{References}
	
	\section*{Supplementary materials}
	\markboth{Supplementary materials}{}
	\newcommand{\beginsupplement}{%
		\setcounter{table}{0}
		\renewcommand{\thetable}{S\arabic{table}}%
		\setcounter{figure}{0}
		\renewcommand{\thefigure}{S\arabic{figure}}%
	}

	\beginsupplement
	\begin{table}[H]
		\centering
		\caption{Mean IoU values for different training pipelines (columns, P1/P2/P3/P4) and different ratios of the labeled data (rows) in the semi-supervised approach.
			\label{t3}} \vspace*{5mm}

		\begin{tabular}{ m{1.7cm}| m{2.7cm} m{2.7cm} m{2.7cm} m{2.7cm} }
			\cline{2-5} 
			\multirow{3}{*}{ } & \multicolumn{4}{c}{\textbf{PhC dataset}} \\ \cline{2-5}
			& \multicolumn{2}{c}{KL loss}&\multicolumn{2}{c}{IID loss } \\ \cline{2-5}
			& Best supervised loss (P1)&	Best final loss (P2) &	Best supervised loss (P3)&Best final loss (P4) \\ \hline
			10\% ratio&	0.9410&	0.9380&	0.8841 & 0.9252\\ \hline
			25\% ratio&	0.9295	&0.9416&	0.9303 & 0.8635 \\ \hline
			50\% ratio&	0.9358&	0.9192&	0.9436 & 0.9418 \\ \hline \cline{2-5}	
			
			\cline{2-5} 
			\multirow{3}{*}{ } & \multicolumn{4}{c}{\textbf{FluoGFP dataset}} \\ \cline{2-5}
			& \multicolumn{2}{c}{KL loss}&\multicolumn{2}{c}{IID loss } \\ \cline{2-5}
			& Best supervised loss (P1)&	Best final loss (P2) &	Best supervised loss (P3)&Best final loss (P4)  \\ \hline
			10\% ratio&0.8777&	0.8788&	0.8727&	0.8364\\ \hline
			25\% ratio&	0.9189&	0.8525&	0.9308&	0.8482 \\ \hline
			50\% ratio&	0.9281&	0.8844&	0.9150&	0.9083 \\ \hline \cline{2-5}

			\multirow{3}{*}{ } & \multicolumn{4}{c}{\textbf{BigBrain dataset}} \\ \cline{2-5}
			& \multicolumn{2}{c}{KL loss}&\multicolumn{2}{c}{IID loss } \\ \cline{2-5}
			& Best supervised loss (P1)&	Best final loss (P2) &	Best supervised loss (P3)&Best final loss (P4)  \\ \hline
			10\% ratio&0.6504&	0.6616&	0.6733&	0.6137\\ \hline
			25\% ratio&	0.6991&	0.6591&	0.6895&	0.6643 \\ \hline
			50\% ratio&	0.6805&	0.6887&	0.6927&	0.6862 \\ \hline \cline{2-5}

			\multirow{3}{*}{ } & \multicolumn{4}{c}{\textbf{FluoHoechst dataset}} \\ \cline{2-5}
			& \multicolumn{2}{c}{KL loss}&\multicolumn{2}{c}{IID loss } \\ \cline{2-5}
			& Best supervised loss (P1)&	Best final loss (P2) &	Best supervised loss (P3)&Best final loss (P4)  \\ \hline
			10\% ratio&	0.9273&	0.9273&	0.9516&	0.9350\\ \hline
			25\% ratio&	0.9427&	0.9043&	0.9340&	0.9340\\ \hline
			50\% ratio&	0.9377&	0.8930&	0.9343&	0.8114 \\ \hline

		\end{tabular}
		
	\end{table}
	
		\begin{table}[H]
		\centering
		\caption{Mean IoU values for different datasets, calculated for the output of the main and auxiliary head in the self-supervised approach. For auxiliary output, classes were assigned in accordance to maximally overlapping areas.
			\label{t3}} \vspace*{5mm}

		\begin{tabular}{ m{3cm}| m{3cm} | m{3cm} | m{3 cm} }
			
			& PhC dataset &	FluoGFP dataset  &  FluoHoechst dataset \\ \hline
		 \hline
			auxiliary head  &	0.3475	& 0.6983 &	0.7110 \\ \hline
			main head  &	0.2409	&0.2133&	0.4631 \\
			
		\end{tabular}
		
	\end{table}
	
			
		
	

\end{document}